\documentclass[10pt,twocolumn,letterpaper]{article}

\usepackage[pagenumbers]{cvpr} 

\usepackage{multirow}

\usepackage{lipsum}

\newcommand{\xdeleted}[1]{}

\definecolor{cvprblue}{rgb}{0.21,0.49,0.74}
\usepackage[pagebackref,breaklinks,colorlinks,allcolors=cvprblue]{hyperref}

\def\paperID{} 
\def\confName{CVPR}
\def\confYear{2026}

\title{MedSAD-CLIP: Supervised CLIP with Token-Patch Cross-Attention for Medical Anomaly Detection and Segmentation}
\author{
Thuy Truong Tran$^{1}$ \quad
Minh Kha Do$^{2}$ \quad
Phuc Nguyen Duy$^{3}$ \quad
Min Hun Lee$^{1}$\thanks{Corresponding author} \\[0.5em]
$^{1}$Singapore Management University \quad
$^{2}$La Trobe University \quad
$^{3}$Vietnam National University, Hanoi \\[0.5em]
{\tt\small tt.tran.2024@phdcs.smu.edu.sg \quad
m.do@latrobe.edu.au \quad
phucnd.hp.vn@gmail.com \quad
mhlee@smu.edu.sg}
}

\begin{document}
\maketitle

\newcommand{\mymodel}{MedSAD-CLIP\xspace}
\newcommand{\myloss}{MC-Loss\xspace}
\newcommand{\mylossfull}{Margin-based image-text Contrastive Loss\xspace}
\newcommand{\myAtt}{TPCA\xspace}
\newcommand{\myAttfull}{Token-Patch Cross-Attention\xspace}

\begin{abstract}
Medical anomaly detection (MAD) and segmentation play a critical role in assisting clinical diagnosis by identifying abnormal regions in medical images and localizing pathological regions.
Recent CLIP-based studies are promising for anomaly detection in zero-/few-shot settings, and typically rely on global representations and weak supervision, often producing coarse localization and limited segmentation quality.
In this work, we study supervised adaptation of CLIP for MAD under a realistic clinical setting where a limited yet meaningful amount of labeled abnormal data is available.
Our model \mymodel leverages fine-grained text-visual cues via the \myAttfull(\myAtt) to improve lesion localization while preserving the generalization capability of CLIP representations. Lightweight image adapters and learnable prompt tokens efficiently adapt the pretrained CLIP encoder to the medical domain while preserving its rich semantic alignment. Furthermore, a \mylossfull is designed to enhance global feature discrimination between normal and abnormal representations.
Extensive experiments on four diverse benchmarks-Brain, Retina, Lung, and Breast datasets-demonstrate the effectiveness of our approach, achieving superior performance in both pixel-level segmentation and image-level classification over state-of-the-art methods. Our results highlight the potential of supervised CLIP adaptation as a unified and scalable paradigm for medical anomaly understanding. Code will be made available at \url{https://github.com/thuy4tbn99/MedSAD-CLIP}
\end{abstract}


\section{Introduction}
\label{sec:intro}

\begin{figure}[t]
  \centering
  \includegraphics[width=\columnwidth]{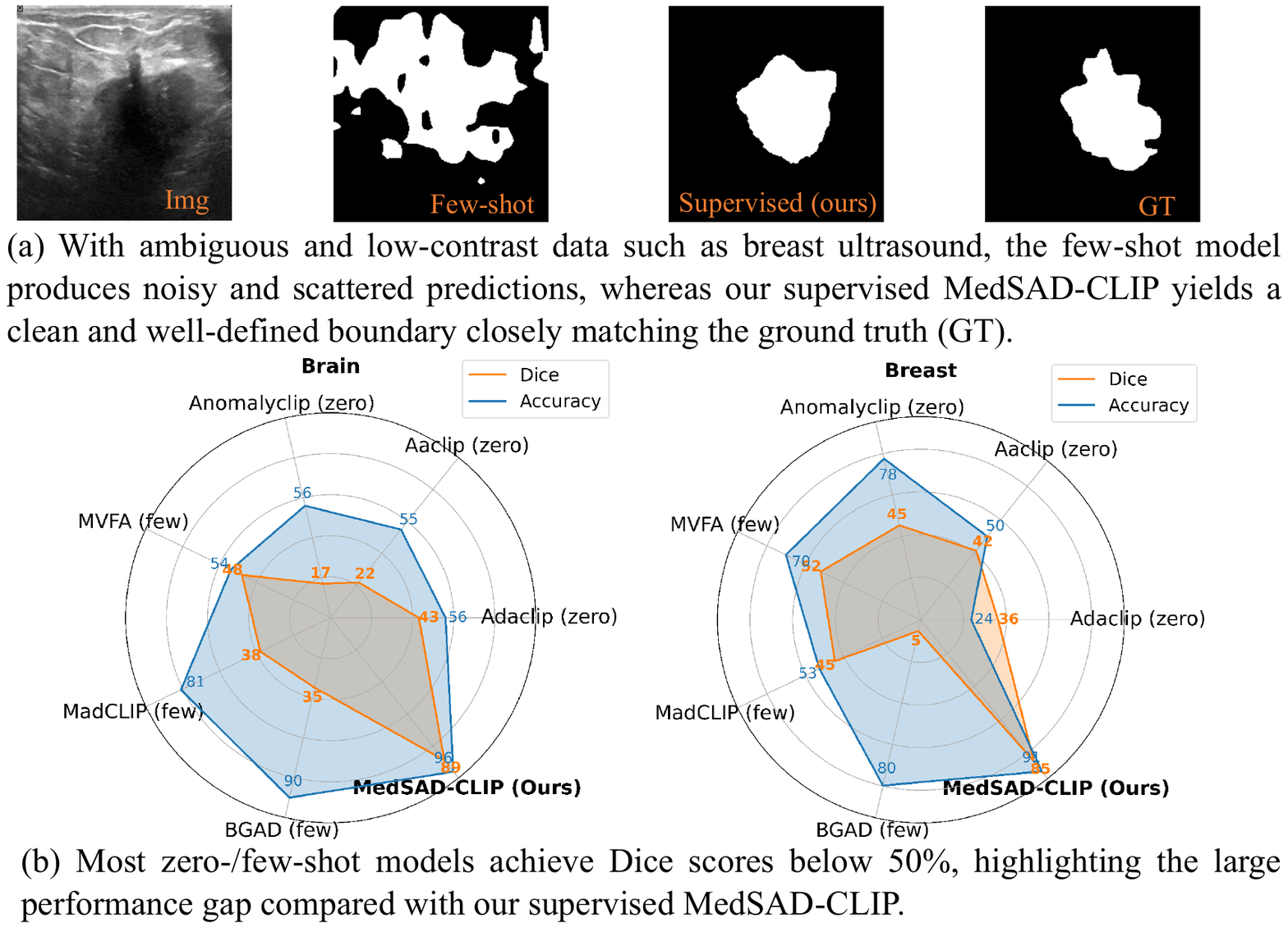}
  \caption{(a) Qualitative result comparisons of few-shot MVFA and our supervised MedSAD-CLIP on a Breast image. (b) Radar comparison of MedSAD-CLIP and zero-/few-shot CLIP-based SOTAs on Brain and Breast datasets.}
  \label{fig:radar_intro}
\end{figure}

Anomaly detection (AD) plays an important role in medical image analysis, providing critical insights that assist clinicians in achieving accurate and timely diagnoses \cite{lee2025ai}. However, medical AD remains challenging due to the ambiguous definition of abnormality and the diverse manifestations of pathological patterns across modalities. In practice, AD must support two coupled outcomes (i.e., image-level anomaly detection and precise localization of lesions).
Recently, zero-/few-shot methods \cite{jeong2023winclip, zhou2023anomalyclip, gong2025fe, ma2025aa, cao2024adaclip} leverage pre-trained vision-language models (VLMs) with carefully designed textual prompts and adaptation schemes to enhance image-level anomaly detection. While these approaches appeal due to their scalability, they may yield coarse segmentation maps with low overlap with ground-truth lesions and exhibit poor calibration across domains, limiting reliability in practice (as shown in \cref{fig:radar_intro} with low Dice score). 
Importantly, this limitation is not solely attributed to the zero-/few-shot setting itself. Rather, it is closely related to the inherent design of CLIP, where the dominance of global representations tends to weaken the semantic correspondence between individual image patches and textual descriptions \cite{shao2024explore}.

Despite the widespread adoption of CLIP in zero-/few-shot anomaly detection, limited attention has been given to supervised CLIP-based frameworks for medical anomaly detection and segmentation. In many realistic clinical scenarios, a limited yet meaningful amount of labeled abnormal data is available, providing an opportunity to further exploit CLIP representations for more reliable localization while preserving the generalization benefits of vision-language pretraining.
To better understand the role of supervision in CLIP-based medical anomaly analysis, we study supervised CLIP adaptation and compare it with existing zero-/few-shot VLM approaches under realistic clinical settings where a limited yet meaningful amount of labeled abnormal data is available.
Firstly, \myAttfull is employed to establish fine-grained, direct interactions between individual text tokens and image patches, replacing the reliance on a global sentence embedding and enabling sharper lesion boundary extraction. Unlike existing work using dot-product heatmaps directly as masks, \myAtt forms token-conditioned features and concatenates them with rich image features before decoding, injecting granular language cues while preserving visual detail and sharpening lesion boundaries.
Secondly, we integrate multi-level image adapters and a learnable text prompt module into the pretrained CLIP \cite{huang2024adapting,zhang2024mediclip}. Image adapters transfer general-domain representations to medical imaging, while the learnable prompt module further facilitates adaptation to various clinical contexts, anatomical regions, and imaging modalities. 
Lastly, we propose \mylossfull (\myloss), which explicitly enforces a larger semantic distance between normal and abnormal image-text pairs, guiding the model to learn a discriminative boundary and thereby improving image-level anomaly detection. Moreover, this loss enhances both the image adaptation module and the learnable prompt tuning, enabling them to align visual-textual representations more effectively. 

In summary, our main contributions are:
\begin{itemize}
    \item We empirically quantify and reveal a significant performance gap within the CLIP-based paradigm itself, comparing zero-/few-shot models against fully supervised counterparts in medical anomaly detection and segmentation. We therefore explore a supervised CLIP-based framework designed to enhance abnormal region segmentation across diverse datasets.
    \item To better exploit fine-grained visual-text information beyond the global representations typically emphasized in CLIP-based models, we introduce a \myAttfull (\myAtt) module that sharpens lesion boundaries and improves segmentation, together with a Margin-based Contrastive Loss (\myloss) to enhance discrimination in ambiguous cases.
    \item Our \mymodel achieves state-of-the-art performance across multiple medical anomaly detection benchmarks, showing robust and consistent generalization.
\end{itemize}

\section{Related Work}
\label{sec:related_work}
\begin{figure*}[t]
  \centering
  \includegraphics[width=0.95\textwidth]{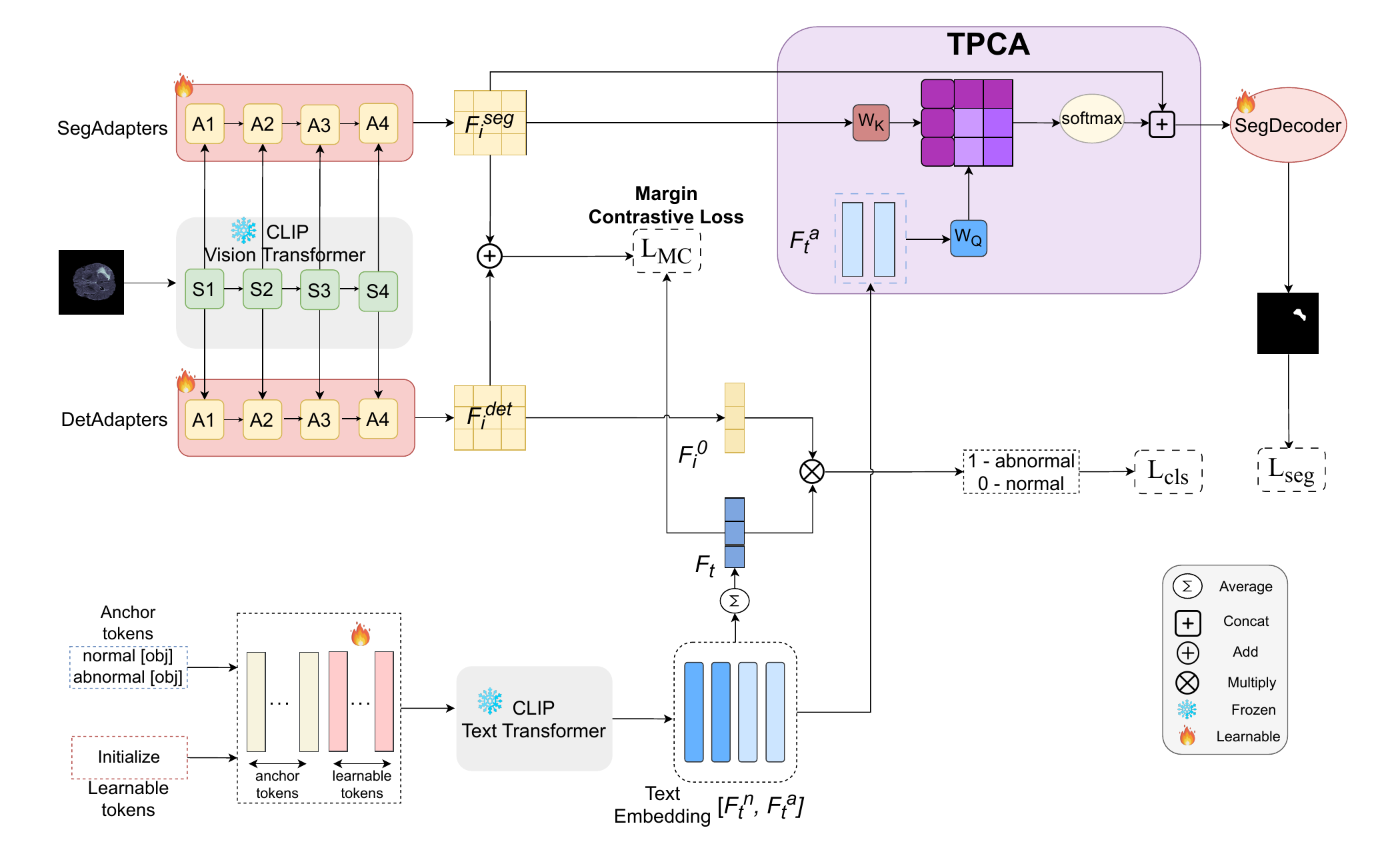}
  \caption{Overall architecture of our proposed \mymodel. It consists of two branches: the classification branch (Det) and the segmentation branch (Seg). The model extends CLIP~\cite{radford2021learning} with adapter modules and learnable prompts to adapt to medical image analysis.}
  \label{fig:model}
\end{figure*}
\subsection{Medical Anomaly Detection}
Existing medical anomaly detection approaches can be broadly categorized into unsupervised and supervised methods.

Unsupervised methods are widely adopted in the medical field, where models are trained exclusively on normal samples and identify anomalies as deviations from the learned distribution of healthy data. Traditional techniques \cite{liu2024rolling, zhou2020encoding, zimmerer2018context, chen2020mama, zong2018deep} often use reconstruction to learn prior information from normal data.  For example, WeakAnD \cite{seebock2019exploiting} leverages Bayes U-Net \cite{saidu2021active}, which is trained on healthy data to model epistemic uncertainty as an indicator of anatomical deviations, while Han et al. proposed MADGAN \cite{han2021madgan}, based on GAN \cite{goodfellow2020generative} with adjacent slices, which enables better exploitation of 3D data to classify anomalies on brain MRI data. More recently, SQUID \cite{xiang2023squid} introduces a teacher–student framework with a memory queue for in-painting-based AD, while DDPM \cite{wolleb2022diffusion} and AnoDDPM \cite{wyatt2022anoddpm} employ denoising diffusion strategies to reconstruct healthy-looking tissue while preserving normal anatomical details.

However, in real-world medical scenarios, a non-trivial portion of anomalous data is typically available during dataset collection \cite{bao2024bmad}. Effectively leveraging these abnormal cases can provide valuable supervisory signals, enabling the model to enhance its predictive capability beyond merely learning from normal patterns. DRA \cite{ding2022catching} learns disentangled representations of both seen and unseen abnormalities, while BGAD \cite{yao2023explicit} introduces a supervised anomaly detection framework that employs an explicit boundary–guided semi–push–pull contrastive learning mechanism to enhance feature discriminability. In contrast, we cast supervised anomaly understanding in a vision–language model and show how token-level textual cues and margin-based calibration jointly improve localization and decision stability.

\subsection{Medical Vision Language Model}
Recently, a growing line of research \cite{zhao2025clip, hartsock2024vision, lee2025ai, liu2023clip} has adapted VLMs \cite{jia2021scaling, radford2021learning, zhang2024vision} to the medical domain, aiming to leverage their powerful vision–language alignment for cross-modal understanding of medical images. LViT \cite{li2023lvit} integrates short disease descriptions with CT or X-ray image information through a lightweight embedding-based text representation and a hybrid CNN–Transformer architecture with pixel-level attention for segmentation. When using pre-trained VLMs like CLIP, it is essential to mitigate the domain gap and adapt or fine-tune the model for medical anomaly detection tasks. For example, MVFA \cite{huang2024adapting}, Aa-CLIP \cite{ma2025aa}, and Fe-CLIP \cite{gong2025fe} leverage vision adapters to transfer knowledge from the general domain to the medical domain, enabling more effective learning of domain-specific visual representations for accurate detection of medical anomalies. However, these models rely on fixed, handcrafted prompts to represent medical concepts, limiting adaptability to diverse imaging contexts and textual variations. MediCLIP \cite{zhang2024mediclip}, AnomalyCLIP \cite{zhou2023anomalyclip}, and MadCLIP \cite{shiri2025madclip} further extend this paradigm by introducing learnable prompt representations that enhance cross-modal alignment and better capture semantic variations in medical anomaly detection. Nevertheless, these methods either lack dedicated vision adapters for image representation or rely on overly simplistic fusion strategies that combine only global textual information with image features, thereby limiting fine-grained cross-modal interaction and hindering precise segmentation performance. In contrast, our approach integrates vision adapters with learnable textual representations and employs token-patch cross-attention that directly aligns pathology text tokens to spatial patches, yielding sharper segmentation boundaries. Most CLIP-based anomaly/localization methods compute a patch-wise similarity
(e.g., cosine/dot product) between image patches and a global text embedding and
\emph{use the similarity map itself} as the segmentation cue \cite{zhou2023anomalyclip,
gong2025fe,ma2025aa}. A few variants expose cross-attention weights but still treat the
weights/score maps as masks, effectively replacing the underlying visual representation.
In contrast, our Token–Patch Cross-Attention (TPCA) is not a dot-product mask:
we form token-conditioned features via attention and \emph{concatenate} them with the
original image features before decoding. This preserves low-level texture/shape
information in the visual stream while injecting fine-grained, token-level cues as an
add-on signal, yielding sharper boundaries without discarding visual detail.

\subsection{Margin Contrastive loss}
Several works enhance contrastive learning by introducing explicit margins in the similarity space.
Shah et al.\ propose Max-Margin Contrastive Learning (MMCL)~\cite{shah2022max} which injects a margin to enlarge the separation between positives and hard negatives, while
Reiss and Hoshen~\cite{reiss2023mean} introduce the Mean-Shifted Contrastive Loss that shifts features by mean to change the geometry of the contrastive objective to better separate normal vs anomaly. 
In the medical domain, supervised contrastive learning with an angular margin has been applied to diabetic retinopathy detection~\cite{zhusupervised}, where an additional angular gap between classes yields more discriminative retinal images.
In contrast, we introduce a \mylossfull tailored to supervised medical anomaly detection. This simple hinge-style formulation explicitly pushes image features closer to normal/abnormal text features and away from the opposite category, enforcing an absolute similarity gap between abnormal and normal prototypes, yielding more calibrated, accurate image-level anomaly detection.

\section{Methodology}

\subsection{Problem Formulation}
Our objective is to design a supervised anomaly detection model that jointly performs image-level anomaly detection as a binary classification and pixel-level segmentation. Given an input image $\mathbf{I} \in \mathbb{R}^{H \times W \times 3}$, where ${H}$ and ${W}$ denote the image height and width, respectively, the model aims to produce an anomaly score $S \in [0,1]$ indicating whether the image is normal or abnormal, and a segmentation map ${G} \in \mathbb{R}^{H \times W}$ highlighting pixel-wise abnormal regions. Unlike the unsupervised AD setting, the supervised training set includes both normal and abnormal samples, formally defined as $D=D^n \cup D^a$ where $D^n$ and $D^a$ denote the sets of normal and abnormal images, respectively. The goal is summarized as a parametric function $\mathcal{F}_{\theta}: \mathbf{I} \mapsto (S, {G})$. 

\subsection{Overall Architecture}
\cref{fig:model} illustrates the overall architecture of our proposed model \mymodel. The model takes two inputs: a medical image and a text prompt.
We build upon the CLIP framework \cite{radford2021learning}, which aligns image and text representations through a Vision Transformer and Text Transformer. To adapt to medical domain, we introduce adapter modules, inspired by the MVFA model \cite{huang2024adapting}, to extract domain-specific semantic features, and learnable prompt embeddings, following the prompt-learning strategy in \cite{zhang2024mediclip}, to adaptively generalize across various medical anomaly detection tasks. Specifically, after passing the image through CLIP Vision Transformer, our framework incorporates two dedicated adapters: the {DetAdapter} for classification and the {SegAdapter} for segmentation. In parallel, the learnable text prompt is fed into the Text Transformer to obtain text embedding. For classification, we multiply the first image patch token representing global feature and the average normal/abnormal text embedding to obtain anomaly score. To enhance spatially precision segmentation, we introduce the \myAttfull{} module (\myAtt), which performs fine-grained attention between each text token and each image patch, enabling more accurate cross-modal alignment. The attention features are then concatenated with image features from the {SegAdapter} via a residual connection~\cite{he2016deep}. This design preserves low-level spatial details while enriching the fused representation with fine-grained semantic cues. The resulting feature map is then forwarded to the {SegDecoder} to produce the final segmentation mask.
Finally, we propose \mylossfull with a fixed cosine similarity to enhance the separability between normal and abnormal global representations. Particularly, for normal images, the similarity between image and normal text features is encouraged to be higher than that with abnormal text features by at least a fixed margin, whereas for abnormal images, the reverse relation is enforced. This mechanism enables the model to more effectively distinguish between normal and abnormal cases.

\subsection{Multi-level Image Adapters and Learnable Prompt}

\textbf{Multi-level Image Adapters} Although CLIP demonstrates strong generalization on open-domain datasets, its representations are not optimized for the specialized characteristics of medical images. To bridge this domain gap, we integrate a {Multi-level Image Adaptation} strategy that employs lightweight adapter modules to shift the pretrained CLIP visual encoder into the medical anomaly detection domain. The proposed module is divided into two branches: DetAdapter, designed for image-level classification, and SegAdapter, tailored for pixel-level segmentation.

Given an input image 
\(
\mathbf{I} \in \mathbb{R}^{H \times W \times 3},
\)
the CLIP image encoder \( f_{\text{clip}}(\cdot) \) first extracts hierarchical patch embeddings:
\begin{equation}
    F_0 = f_{\text{clip}}(\mathbf{I}) \in \mathbb{R}^{N_i \times D},
\end{equation}
where \( N_{i} \) is the number of patches, and \( D \) denotes the feature dimension. To adapt these representations to medical data, we insert four sequential adapter stages into the encoder pipeline. Each adapter $A_k$ with $k=1,2,3,4$ consists of two linear projection layers followed by normalization and nonlinearity, defined as:
\begin{equation}
    A_{k}(F_{0}) = \phi^2_{k}(\text{LN}^2(W_{k}^2 \, \phi^1_{k}(\text{LN}^1(W_{k}^1 F_0))) \big), 
\end{equation}

where \( \text{LN}(\cdot) \) denotes layer normalization, \( W \) represents a learnable linear projection, and \( \phi\) is a LeakyReLU \cite{maas2013rectifier} activation function. We then average the outputs to obtain the adapted feature:
\begin{equation}
\mathbf{F}_i = \frac{1}{4} \sum_{k=1}^{4}A_{k}(F_{0})
\end{equation}

\noindent \textbf{Learnable Prompt}. To ensure adaptability across diverse diseases and anatomical regions, we incorporate a learnable prompt mechanism, which not only contains normal/abnormal semantic guidance but also learns to 
flexibly represent contextual information. Our design is motivated by recent vision-language adaptation frameworks~\cite{zhang2024mediclip, jeong2023winclip}, which demonstrate that flexible text conditioning can significantly enhance cross-domain generalization.
Specifically, we initialize two anchor templates to represent the normal and abnormal concepts:
\begin{align}
T_{\text{norm}} &= \text{``a photo of a normal [obj]''}, \\
T_{\text{abn}} &= \text{``a photo of a damaged [obj]''},
\end{align}
where \([obj]\) denotes the medical category, such as \emph{brain} or \emph{retina}. 
To improve their adaptability, we append a set of $K$ learnable tokens $\{p_1, p_2, \dots, p_K\}$ to the end of each anchor prompt, forming the learnable prompt representations:
\begin{align}
\tilde{T}_{\text{norm}} = [T_{\text{norm}}, p_1, \dots, p_K], \\
\tilde{T}_{\text{abn}} = [T_{\text{abn}}, p_1, \dots, p_K].
\end{align}

Unlike previous studies \cite{zhang2024mediclip, huang2024adapting} that average embeddings across multiple handcrafted prompt variations, we empirically observe that these learnable tokens are sufficient to capture the semantic variability between normal and abnormal states. This design allows the model to automatically adjust to different organs or disease contexts without manually crafting diverse prompt templates.
Then, we encode them through the CLIP text encoder to obtain semantic text embeddings:
\begin{equation}
\mathbf{F}_t = \text{TextEncoder}([\tilde{T}_{\text{norm}}, \tilde{T}_{\text{abn}}]),
\end{equation}
where $\mathbf{F}_t \in \mathbb{R}^{2 \times N_t \times D}$ denotes the dual text feature representation for the normal and abnormal descriptions, and $D$ is the embedding dimension of the shared CLIP space, $N_t$ denotes the fixed length of the prompt.

\subsection{\myAttfull for Segmentation}

In this section, we introduce a \myAttfull (\myAtt) module, inspired by recent vision–language interaction methods \cite{xu2023multimodal}. 
Unlike existing CLIP-based anomaly detection models that employ simple fusion techniques such as concatenation \cite{huang2024adapting, jeong2023winclip, zhang2024mediclip, vu2024bm} or average aggregation \cite{cao2024adaclip}  before generating the anomaly map, our \myAtt explicitly models the fine-grained correspondence between textual semantics and localized image regions. Using cross-attention, we obtain token-conditioned feature maps, which are then fused with the original image representations before decoding. This integration retains low-level structural detail while introducing fine-grained, token-level semantics as complementary guidance.

Specifically, the \myAtt module receives two inputs: the abnormal prompt embeddings $\mathbf{F}_{t}^{a} \in \mathbb{R}^{N_t \times D}$, obtained from the text encoder, and the spatial image features $\mathbf{F}_i^{s} \in \mathbb{R}^{N_i \times D}$, extracted from the segmentation adapter branch. 

To align the two modalities, we employ a scaled dot-product cross-attention mechanism \cite{vaswani2017attention} that computes only the attention weights corresponding to the abnormal text tokens as:
\begin{equation}
\mathbf{A}_{t \rightarrow i}^{\text{abn}} = 
\mathrm{Softmax}\!\left(
\frac{(\mathbf{F}_{t}^{a} \mathbf{W}_q)(\mathbf{F}_i^{s} \mathbf{W}_k)^{\top}}
{\sqrt{d_k}}
\right),
\label{eq:mca}
\end{equation}
where $\mathbf{W}_q, \mathbf{W}_k \in \mathbb{R}^{D \times d_k}$ are learnable projection matrices. To be more specific, each text token is treated as a semantic query, while the image patch embeddings serve as key and value. This design allows the attention map to capture the degree of semantic association between textual concepts and corresponding image regions. We employ text tokens as queries rather than image features to localize the image patches that potentially contain abnormal semantics. For example, a text prompt such as ``a photo of a damaged brain`` acts as a semantic query that identifies the regions in the brain image corresponding to the concept of ``damaged``. In anomaly detection, this mechanism is particularly valuable since language inherently defines abnormality context, whereas images merely provide its visual manifestation.

After that, we integrate these attention-derived features with the spatial image patch tokens as a residual connection \cite{he2016deep} to guide segmentation. The multimodal fusion feature can be expressed as:
\begin{equation}
\mathbf{F}_{\text{fuse}} = 
\mathrm{Concat}\!\left(
\mathbf{F}_i^{s},
\, 
\mathrm{Permute} \!\big(
\mathrm{Mean}_h(\mathbf{A}_{t \rightarrow i}^{\text{abn}})
\big)
\right),
\label{eq:fusion}
\end{equation}
where $\mathrm{Mean}_h(\cdot)$ averages the attention across attention heads, and $\mathrm{Permute}(\cdot)$ swaps token–patch dimensions to align spatially with $\mathbf{F}_i^{s}$. 

Finally, the segmentation mask is generated via a lightweight SegDecoder $\mathcal{D}_{\text{seg}}$ followed by bilinear upsampling and softmax normalization:
\begin{equation}
{G} = 
\mathrm{Softmax}\!\Big(
\mathcal{U}\!\big(
\mathcal{D}_{\text{seg}}(\mathbf{F}_{\text{fuse}})
\big)
\Big),
\label{eq:segmask}
\end{equation}
where $\mathcal{U}(\cdot)$ denotes bilinear interpolation to the original image size. The resulting $\mathbf{G}$ represents pixel-wise probabilities for abnormal regions.

Intuitively, this design provides a dual benefit: (1) textual prompts act as semantic anchors that refine spatial localization of anomalies, and (2) image patches contribute localized cues to disambiguate subtle variations within complex medical regions.

\subsection{Classification}
In parallel with the segmentation pathway, we design a classification branch to predict image-level anomaly detection. 
Given the image patch tokens extracted from the DetAdapter, denoted as $\mathbf{F}_i^{d} = \{\mathbf{f}_0, \mathbf{f}_1, \ldots, \mathbf{f}_{N_i}\} \in \mathbb{R}^{(N_i+1) \times D}$, the first token $\mathbf{f}_0$ corresponds to the \texttt{[CLS]} token in CLIP, which encodes the global image representation. Meanwhile, the text encoder produces prompt embeddings for both normal and abnormal descriptions, represented as $\mathbf{F}_t^{n}, \mathbf{F}_t^{a} \in \mathbb{R}^{N_t \times D}$. 
We compute their mean representations to obtain compact semantic prototypes:
\begin{equation}
\bar{\mathbf{t}}_{n} = \frac{1}{N_t} \sum_{j=1}^{N_t} \mathbf{F}_{t,j}^{n}, 
\qquad 
\bar{\mathbf{t}}_{a} = \frac{1}{N_t} \sum_{j=1}^{N_t} \mathbf{F}_{t,j}^{a}.
\label{eq:text_mean}
\end{equation}
The anomaly classification score is computed by projecting the global image token onto the text feature space through an inner product:
\begin{equation}
S = \sigma(\mathbf{f}_0 \mathbf{T}^{\top}), 
\qquad 
\mathbf{T} = [\,\bar{\mathbf{t}}_{n},\, \bar{\mathbf{t}}_{a}\,]^{\top} \in \mathbb{R}^{2 \times D},
\label{eq:dot_cls}
\end{equation}
where $\sigma(\cdot)$ denotes the softmax activation. 
A higher $S$ indicates that the image is more semantically similar to the abnormal description.

\subsection{Loss function}
The anomaly detection can be seen as a binary classification task. The model predicts an anomaly score 
$S \in [0,1]$ for each input image $\mathbf{I}$. 
We assign a binary ground-truth label $y \in \{0,1\}$, 
where $y = 1$ indicates an {abnormal} image and $y = 0$ indicates a {normal} one. 
The classification objective is to minimize the Binary Cross-Entropy (BCE) loss between 
the predicted score $S$ and the ground-truth label $y$:
\begin{equation}
\mathcal{L}_{cls} = - \frac{1}{N} \sum_{i=1}^{N} 
\Big[ y_i \log S_i + (1 - y_i) \log (1 - S_i) \Big],
\label{eq:bce_loss}
\end{equation}
where $S_i$ denotes the predicted anomaly score for the $i$-th image, $y_i$ is its corresponding ground-truth label and $\log$ denotes the logarithm operation. 
Similarly, the model predicts a segmentation map  $G \in [0,1]^{H \times W}$ for each input image $\mathbf{I}$, 
with ground-truth mask ${M} \in \{0,1\}^{H \times W}$.
To achieve accurate localization under severe class imbalance, we combine the Dice loss \cite{sudre2017generalised} and the Focal loss \cite{lin2017focal}: 

\begin{equation}
\mathcal{L}_{seg} = 
\frac{1}{N} \sum_{i=1}^{N} 
\Big[ 
\text{Focal} (G_i, M_i) 
+ 
\text{Dice}(G_i, M_i)
\Big].
\label{eq:seg_loss}
\end{equation}

Dice loss enhances overlap consistency between predictions and ground truth, while Focal loss focuses learning on hard-to-classify abnormal pixels. 

\noindent \textbf{\mylossfull}. To enhance the alignment between visual and textual representations, 
we introduce a \myloss that explicitly enforces 
a margin-based separation between the similarity of image-normal and image-abnormal text pairs. 
The loss is defined using image feature $\mathbf{F}_i \in \mathbb{R}^D$, the corresponding text embeddings of normal/abnormal descriptions 
$\mathbf{\bar{t}}_n, \mathbf{\bar{t}}_a \in \mathbb{R}^D$, and factor $\bar{y_i}=1$ for normal and $\bar{y_i}=-1$ for abnormal. The cosine similarities are computed as:
\[
s^+_i = \cos(\mathbf{f}_i, \mathbf{\bar{t}}_n), \quad 
s^-_i = \cos(\mathbf{f}_i, \mathbf{\bar{t}}_a),
\]
The \myloss is formulated as:
\begin{equation}
\mathcal{L}_{MC} = 
\frac{1}{N} \sum_{i=1}^{N} 
\max \big(0, \, \tau - \bar{y_i} (s^+_i - s^-_i) \big),
\label{eq:mcl}
\end{equation}
where $\tau > 0$ is a predefined margin hyperparameter. 
For normal samples ($\bar{y}_i = +1$), the loss encourages 
$s^+_i > s^-_i + \tau$, meaning that the image embedding should be 
closer to the \textit{normal} text embedding than to the abnormal one by at least a margin $\tau$. 
Conversely, for abnormal samples ($\bar{y}_i = -1$), the constraint is reversed, enforcing 
$s^-_i > s^+_i + \tau$.
This \mylossfull design effectively pushes image embeddings toward semantically aligned textual representations while pushing them away from opposite-category descriptions beyond the margin.  By minimizing $\mathcal{L}_{MC}$, the model learns more discriminative multimodal representations, resulting in improved visual-textual alignment for anomaly classification and segmentation. By introducing an explicit cosine margin, the model learns fine-grained semantic separation crucial for detecting ambiguous abnormalities.
Finally, the overall loss is as followed:
\begin{equation}
\mathcal{L} = 
\mathcal{L}_{cls} + \mathcal{L}_{seg} + \mathcal{L}_{MC}
\label{eq:loss_all}
\end{equation}

\section{Experiments}

\subsection{Datasets and Metrics}

\begin{table}[t]
\centering
\caption{Summary of datasets used.}
\resizebox{\columnwidth}{!}{
\begin{tabular}{lcccc}
\toprule
\textbf{Dataset} & \textbf{Anatomical region} & \textbf{Image type} & \textbf{\#Train} & \textbf{\#Test} \\
\midrule
Brain        & Brain Tumor     & MRI         & 3715 & 83  \\
Retina   & Retinal Edema   & OCT         & 1805 & 115 \\
Lung & Lung Covid      & CT          & 894  & 383 \\
Breast       & Breast Cancer   & Ultrasound  & 486  & 161 \\
\bottomrule
\end{tabular}}
\label{tab:dataset_summary}
\end{table}

\begin{table*}[t]
\centering
\caption{Comparison results of \mymodel and baselines under {Zero-shot}, {Few-shot}, and {Supervised} settings on four medical anomaly detection datasets. Dice (\%) and Accuracy (\%) are reported. The best results are in \textbf{bold}, and the second best are \underline{underlined}.}
\label{tab:experiment_results}
\resizebox{1.\textwidth}{!}{
\begin{tabular}{llcccccccc}
\toprule
\multirow{2}{*}{\textbf{Setting}} & \multirow{2}{*}{\textbf{Method}} & \multicolumn{2}{c}{\textbf{Brain}} & \multicolumn{2}{c}{\textbf{Retina}} & \multicolumn{2}{c}{\textbf{Lung}} & \multicolumn{2}{c}{\textbf{Breast}} \\
\cmidrule(lr){3-4}\cmidrule(lr){5-6}\cmidrule(lr){7-8}\cmidrule(lr){9-10}
& & \textit{Dice} & \textit{Acc} & \textit{Dice} & \textit{Acc} & \textit{Dice} & \textit{Acc} & \textit{Dice} & \textit{Acc} \\ 
\midrule
\multirow{3}{*}{Zero-shot} 
& Adaclip\cite{cao2024adaclip} & 43.52 & 56.63 & 38.62 & 61.74 & 9.09 & 56.25 & 36.19 & 23.60 \\
& Aaclip\cite{ma2025aa} & 22.68 & 55.42 & 47.98 & 73.91 & 21.73 & 66.41 & 41.54 & 50.00 \\
& Anomalyclip\cite{zhou2023anomalyclip} & 43.25 & 53.01 & 37.25 & 60.86 & 7.47 & 55.46 & 13.52 & 26.92 \\
\midrule
\multirow{3}{*}{Few-shot} 
& MVFA\cite{huang2024adapting} & 48.34 & 54.87 & 66.93 & 79.82 & 68.54 & 76.56 & 52.09 & 70.22 \\
& MadCLIP\cite{shiri2025madclip} & 38.55 & 81.93 & 85.22 & 83.48 & 63.84 & 92.19 & 44.71 & 52.81 \\
& BGAD\cite{yao2023explicit} & 35.66 & 90.36 & 44.58 & 97.03 & 15.17 & 76.56 & 5.39 & 79.78 \\
\midrule
\multirow{7}{*}{Supervised} 
& Adaclip\cite{cao2024adaclip} & 46.99 & 48.19 & 89.47 & 96.52 & 84.93 & 92.19 & \underline{79.87} & 87.64 \\
& Aaclip\cite{ma2025aa} & 44.10 & 85.54 & 73.00 & 95.65 & 77.85 & 96.88 & 8.53 & 23.60 \\
& Anomalyclip\cite{zhou2023anomalyclip} & 31.85 & 53.01 & 34.26 & 80.86 & 20.38 & 74.21 & 16.07 & 82.69 \\
& MVFA\cite{huang2024adapting} & 64.71 & 92.68 & 91.18 & 97.12 & 73.19 & 94.53 & 76.57 & \underline{88.76} \\
& MadCLIP\cite{shiri2025madclip} & 52.70 & 93.98 & 93.07 & 95.65 & 80.22 & \underline{99.22} & 59.58 & 87.08 \\
& BGAD\cite{yao2023explicit} & 68.91 & \underline{95.18} & 63.82 & \underline{97.31} & 41.47 & 91.41 & 5.97 & 64.04 \\
& nnUnet\cite{isensee2021nnu} & \underline{85.94} & -- & 91.12 & -- & \textbf{89.51} & -- & 54.60 & -- \\
& rollingUnet\cite{liu2024rolling} & 67.49 & -- & \textbf{94.61} & -- & 83.69 & -- & 58.01 & -- \\
& \textbf{Ours} & \textbf{89.47} & \textbf{96.34} & \underline{93.18} & \textbf{97.37} & \underline{87.16} & \textbf{99.92} & \textbf{84.96} & \textbf{91.01} \\
\bottomrule
\end{tabular}
}
\end{table*}

In this work, we focus on supervised medical anomaly detection. Therefore, we select datasets that provide both normal and abnormal samples along with corresponding segmentation anomaly maps. Following these criteria, we employ four datasets covering distinct anatomical regions and imaging modalities: Brain~\cite{baid2021rsna, bao2024bmad}, Retina~\cite{hu2019automated, bao2024bmad}, {Lung}~\cite{chang2025unified, fan2020inf}, and {Breast}~\cite{al2020dataset, chang2025unified}. 
For Brain and Retina, the anomaly detection setup follows the BMAD benchmark~\cite{bao2024bmad}, while the preparation protocols for Lung and Breast are based on the SR-ICL framework~\cite{chang2025unified}. The detailed statistics of all datasets are summarized in \cref{tab:dataset_summary}.

\subsection{Experimental Setup}

\noindent \textbf{Evaluation and Implementation.}
We evaluate image-level detection using \textit{Accuracy} and pixel-level segmentation using the \textit{Dice score} \cite{bertels2019optimizing,chicco2020advantages}.
All models are trained on a single NVIDIA L40S GPU with the Adam optimizer ($1\times10^{-4}$ LR, batch size 32, 50 epochs). 
Images are resized to $240\times240$, and we use CLIP ViT-L/14 \cite{radford2021learning} as the backbone with 10 learnable prompt tokens and a fixed margin $\tau{=}0.4$. 
Further training details and ablation configurations are provided in the \textit{Supplementary Material}.

\noindent \textbf{Baselines.}
We compare against recent CLIP-based anomaly detection models under both zero-/few-shot \cite{cao2024adaclip,ma2025aa,zhou2023anomalyclip,huang2024adapting,bao2024bmad,yao2023explicit} and fully supervised settings, with identical preprocessing and partitioning protocols. 
A detailed description of baseline setups is included in the \textit{Supplementary Material}.

\subsection{Main Results}
\cref{tab:experiment_results} summarizes the performance of all baseline methods and our proposed framework across four medical datasets. Overall, our model consistently outperforms both zero-/few-shot and supervised CLIP-based baselines in terms of Dice and Accuracy metrics, when compared with other state-of-the-art CLIP-based methods, such as Aaclip\cite{ma2025aa}, Adaclip\cite{cao2024adaclip}, and MVFA\cite{huang2024adapting}.
We further include strong segmentation baselines, nnU-Net~\cite{isensee2021nnu} and Rolling-Unet~\cite{liu2024rolling}, for a more comprehensive comparison. 
This shows that the \myAttfull Module combined with \mylossfull demonstrates superior capability in distinguishing and localizing anomalies across diverse anatomical regions.

As shown, all three zero-shot models - Adaclip, Aaclip, and Anomalyclip - exhibit weak segmentation performance across all datasets. For example, Adaclip achieves only 9.09\% Dice on the Lung Infection dataset, while Aaclip and Anomalyclip reach merely around 10\% Dice on Breast and Lung images. The remaining results for these models on other datasets also remain below 50\% Dice, indicating poor generalization in zero-shot segmentation.
In addition, their anomaly detection accuracy scores range mostly between 60–80\%, suggesting that while the models can recognize abnormality presence to some extent, they fail to delineate lesion regions precisely.

When limited labeled samples are available, models such as MVFA, MadCLIP, and BGAD show clear improvement over zero-shot methods, particularly in anomaly detection accuracy. For instance, BGAD achieves up to 90.36\% and 97.03\% accuracy on the Brain and Lung Infection datasets, respectively, while MadCLIP obtains 92.19\% on the Lung dataset.
Although the Dice scores also increase compared to the zero-shot setting, the overall segmentation performance remains moderate, generally below 70\%, indicating that a limited number of labeled samples improves detection consistency but still constrains precise lesion localization.

Under the supervised configuration, all baselines benefit from full fine-tuning with labeled data. Methods such as MVFA and MadCLIP exhibit clear improvements, reaching up to 73.19\% Dice and 94.53\% accuracy on the Lung Infection dataset. Despite these gains, their performance remains notably below our proposed framework.
Our method consistently achieves the highest results across all datasets, obtaining 89.47\%, 93.18\%, and 84.96\% Dice on the Brain, Retina, Breast datasets, respectively and second best with 87.16\% Dice on Lung dataset. This corresponds to an average Dice improvement of +17\% over MVFA and +46\% over Aaclip.
Specifically, on the Breast dataset, prior models such as Aaclip, BGAD, and Anomalyclip fail to localize anomalies effectively, yielding Dice scores of only around 10\%. This poor performance stems from the inherently challenging nature of breast anomalies, which exhibit extremely low contrast and open, indistinct contours-where the lesion intensity closely resembles surrounding tissue and the boundaries fade gradually into the background. In contrast, our model achieves nearly 85\% Dice, highlighting its strong generalization ability even under the most challenging conditions.
In summary, the results in \cref{tab:experiment_results} clearly reveal a substantial performance gap between the zero-/few-shot and fully supervised models. Zero-/few-shot CLIP-based methods show limited capability in precise lesion, and even under the supervised setting, they still struggle to achieve consistently strong segmentation accuracy across different medical imaging modalities. Our proposed model, equipped with the \myAtt module and \myloss, achieves a remarkable improvement in anomaly segmentation on all baselines under the fully supervised setting. Furthermore, when compared with strong U-Net-based segmentation baselines, our model achieves higher Dice scores on the Brain dataset and substantially outperforms these approaches on the Breast dataset, with an improvement of approximately 30\%. On the Retina and Lung datasets, our method remains competitive with these specialized segmentation models. Notably, while nnU-Net and Rolling-Unet focus solely on pixel-level segmentation, our framework simultaneously performs anomaly classification and segmentation through vision–language alignment.

\begin{table}[t]
\centering
\caption{Ablation study results on four datasets. We report Dice and Accuracy (\%) for both classification and segmentation tasks.}
\label{tab:ablation_study}
\resizebox{\columnwidth}{!}{
\begin{tabular}{lcccccccc}
\toprule
\textbf{Model} & \multicolumn{2}{c}{\textbf{Brain}} & \multicolumn{2}{c}{\textbf{Retina}} & \multicolumn{2}{c}{\textbf{Lung}} & \multicolumn{2}{c}{\textbf{Breast}} \\
 & \textbf{Dice} & \textbf{Acc} & \textbf{Dice} & \textbf{Acc} & \textbf{Dice} & \textbf{Acc} & \textbf{Dice} & \textbf{Acc} \\
\midrule
CLIP-adapt \cite{huang2024adapting} & 64.71 & 92.68 & 91.18 & 97.12 & 73.19 & 94.53 & 76.57 & 88.76 \\
CLIP-adapt + learnable prompt & 58.62 & 85.37 & 91.79 & 98.25 & 81.43 & 97.66 & 76.40 & 91.01 \\
CLIP-adapt + learnable prompt + \myAtt & 84.24 & 91.46 & 91.94 & 96.49 & 86.10 & 98.44 & 84.00 & 90.70 \\
\textbf{\begin{tabular}[c]{@{}l@{}}\mymodel\\ (learnable prompt +\myAtt + \myloss)\end{tabular}} & \textbf{89.47} & \textbf{96.34} & \textbf{93.18} & \textbf{97.37} & \textbf{87.16} & \textbf{99.22} & \textbf{84.96} & \textbf{91.01} \\
\bottomrule
\end{tabular}
}
\end{table}

\subsection{Qualitative Results}
\begin{figure}[h]
  \centering
\includegraphics[width=\columnwidth]{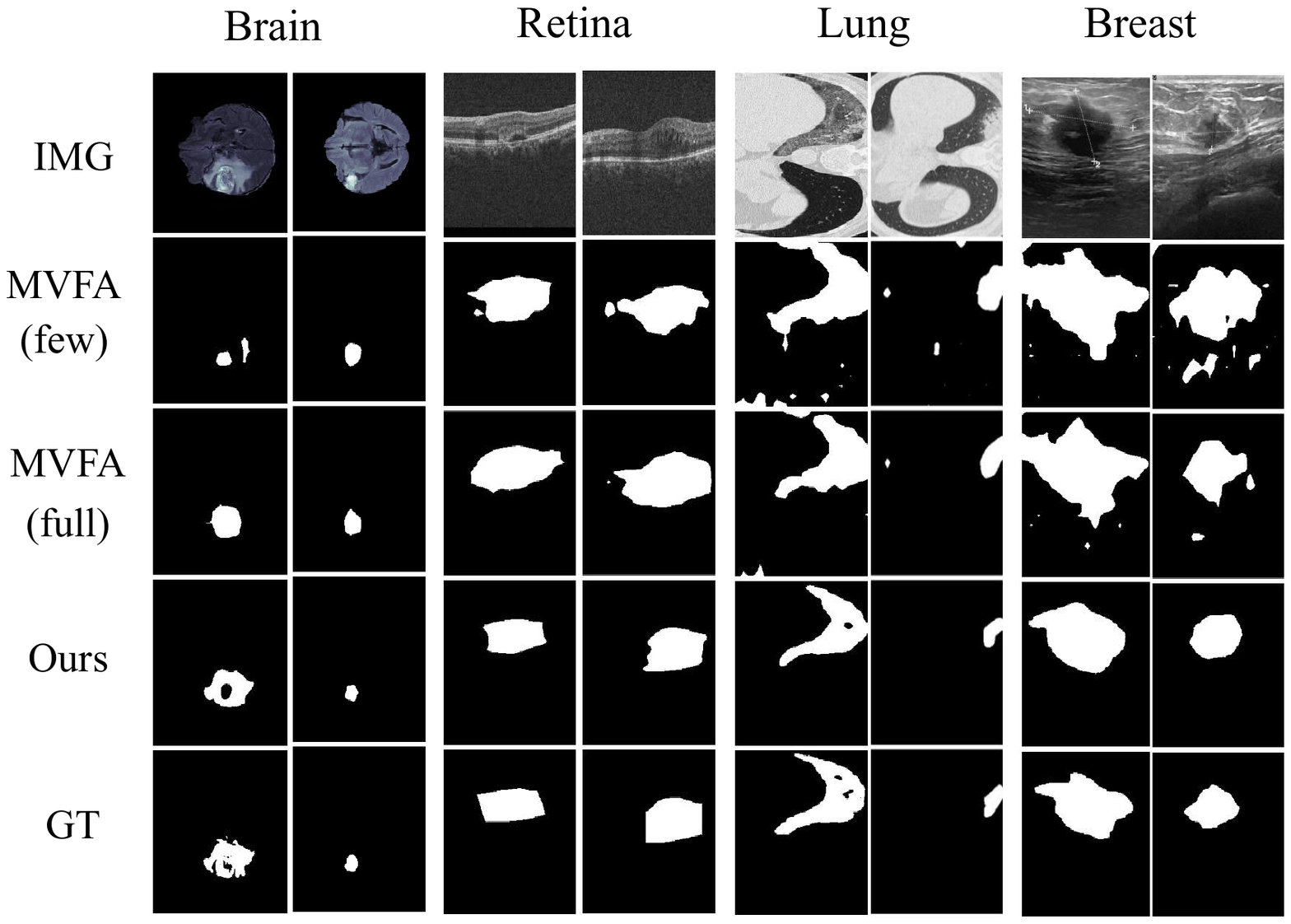}
  \caption{Qualitative comparison between our model MedSAD-CLIP and MVFA. Our MedSAD-CLIP produces sharper and more accurate segmentations.}
  \label{fig:qualitative_result}
\end{figure}

As shown in \cref{fig:qualitative_result}, the MVFA model under both few-shot and fully supervised settings tends to produce noisy or over-segmented predictions, especially in low-contrast regions such as lung and breast images. In particular, the MVFA with few-shot setting often generates wider and less precise masks than its fully supervised counterpart, further highlighting the performance gap between zero-/few-shot and fully supervised models. In contrast, our MedSAD-CLIP yields segmentation maps that closely match the ground-truth shapes and maintain clean boundaries without spurious responses in normal regions. These visual comparisons confirm that the proposed \myAttfull module substantially enhances the model’s ability to localize and delineate abnormal regions more accurately.

\subsection{Ablation Study}
~\cref{tab:ablation_study} presents the impact of each proposed component. Starting from the CLIP-adapt baseline, adding the learnable prompt provides limited improvement, indicating that prompt tuning alone is insufficient for medical anomaly detection. Introducing the \myAtt module yields notable gains across all datasets by enhancing visual–text interaction. In particular, \myAtt plays a crucial role in learning fine-grained segmentation features through the fusion of image patches and text tokens, resulting in a 44\% Dice improvement on the Brain dataset and a 1–3\% gain on others. 
When combined with both \myAtt and \myloss, our full \mymodel achieves the highest Dice and Accuracy scores, confirming the complementary effect of cross-attention and margin-based loss in refining anomaly localization and classification.

\section{Conclusion}
In this paper, we present \mymodel, a supervised CLIP-based framework that significantly advances medical anomaly segmentation. We provide a broad views of limited performance of zero-/few-shot CLIP-based model in segmentation quality due to the nature of global representation.  To leverage fine-grained information, our framework introduces a \myAttfull mechanism and a \mylossfull, enabling more precise abnormality localization and stronger alignment between visual and textual features. In addition, through comparisons with strong U-Net-based baselines, we further demonstrate that our model achieves competitive or superior segmentation performance while simultaneously supporting anomaly classification through vision–language alignment.
{
    \small
    \bibliographystyle{ieeenat_fullname}
    \bibliography{main}
}

\clearpage
\setcounter{page}{1}
\maketitlesupplementary

\section{Experimental Details}
\noindent \textbf{Implementation Details.} 
All experiments are implemented in \textit{PyTorch} \cite{paszke2019pytorch} and conducted on a workstation equipped with a single NVIDIA L40S GPU (48\,GB memory). 
The proposed model is trained using the Adam optimizer \cite{kingma2014adam}  with a learning rate of $1\times10^{-4}$ and a batch size of 32 for 50 epochs. 
Input images are resized to a spatial resolution of $240\times240$ before being fed into the model, and the learnable prompt length is 10 for our experiments. We set the number of learnable tokens to 10, and the margin parameter $\tau$ is fixed at 0.4 by default.
We adopt the CLIP ViT-L/14 \cite{radford2021learning} as the baseline architecture, upon which our adapter and learnable prompt modules are integrated. 
All training and evaluation procedures strictly follow the same preprocessing and data partitioning strategy as in the baseline CLIP configuration to ensure fair comparison. During training, we employ standard data augmentations such as random horizontal flipping and normalization consistent with CLIP preprocessing. 

\noindent \textbf{Baseline Setups.} 
To ensure a fair and comprehensive evaluation, we compare our proposed framework against several strong CLIP-based baselines under two configurations: (1) zero-/few-shot learning and (2) supervised learning.
In the zero-/few-shot configuration, we follow prior medical anomaly detection studies that adapt CLIP pretrained model. Specifically, {Adaclip} \cite{cao2024adaclip}, {Aaclip} \cite{ma2025aa}, and {Anomalyclip} \cite{zhou2023anomalyclip} serve as zero-shot baselines, while  MVFA \cite{huang2024adapting}, MadCLIP \cite{bao2024bmad}, BGAD \cite{yao2023explicit} represent few-shot variants. These models leverage the advantages of CLIP through prompt tuning or lightweight finetuning using either no data or a limited number of samples from the training set.
In contrast, the supervised configuration employs the entire labeled dataset to finetune the CLIP model. For this setup, all the aforementioned baselines are trained in a fully supervised manner. This dual-configuration design enables a systematic comparison between zero-/few-shot adaptation and full supervision, highlighting differences in anomaly detection and segmentation performance across diverse medical datasets.

\section{Comparison of Dice and AUC score}

\begin{table*}[t]
\centering
\small
\caption{Comparison of Dice, Accuracy, and pAUC across four datasets.}
\begin{tabular}{
    p{0.15\textwidth}  
    c c c              
    c c c              
    c c c              
    c c c              
}
\toprule
& \multicolumn{3}{c}{\textbf{Brain}} 
& \multicolumn{3}{c}{\textbf{Retina}}
& \multicolumn{3}{c}{\textbf{Lung}}
& \multicolumn{3}{c}{\textbf{Breast}} \\
\cmidrule(lr){2-4} \cmidrule(lr){5-7} \cmidrule(lr){8-10} \cmidrule(lr){11-13}
\textbf{Method} 
& Dice & Acc & pAUC
& Dice & Acc & pAUC
& Dice & Acc & pAUC
& Dice & Acc & pAUC \\
\midrule
Adaclip (zero) \cite{cao2024adaclip}    
& 43.52 & 56.63 & 90.89  
& 38.62 & 61.74 & 92.92  
& 9.09  & 56.25 & 60.76  
& 36.19 & 23.60 & 86.53 \\
Adaclip (full) \cite{cao2024adaclip}    
& 46.99 & 48.19 & 98.66  
& 89.47 & 96.52 & \textbf{99.68}  
& 84.93 & 92.19 & \textbf{99.63}  
& 79.87 & 87.64 & \textbf{93.18} \\
MVFA (zero) \cite{huang2024adapting}     
& 48.34 & 54.87 & 95.06  
& 66.93 & 79.82 & 97.10  
& 68.54 & 76.56 & 97.19  
& 52.09 & 70.22 & 89.51 \\
MVFA (full) \cite{huang2024adapting}        
& 64.71 & 92.68 & 99.18  
& 91.18 & 97.12 & 99.53  
& 73.19 & 94.53 & 98.07  
& 76.57 & 88.76 & 90.79 \\
\textbf{MedSAD-CLIP}
& \textbf{89.47} & \textbf{96.34} & \textbf{99.54}
& \textbf{93.18} & \textbf{97.37} & {98.97}
& \textbf{87.16} & \textbf{99.22} & {98.47}
& \textbf{84.96} & \textbf{91.01} & {87.48} \\
\bottomrule
\end{tabular}
\label{tab:dice_auc_quantitative}
\end{table*}

\begin{figure*}[h]
  \centering
  \includegraphics[width=0.95\textwidth, height=18cm, keepaspectratio]{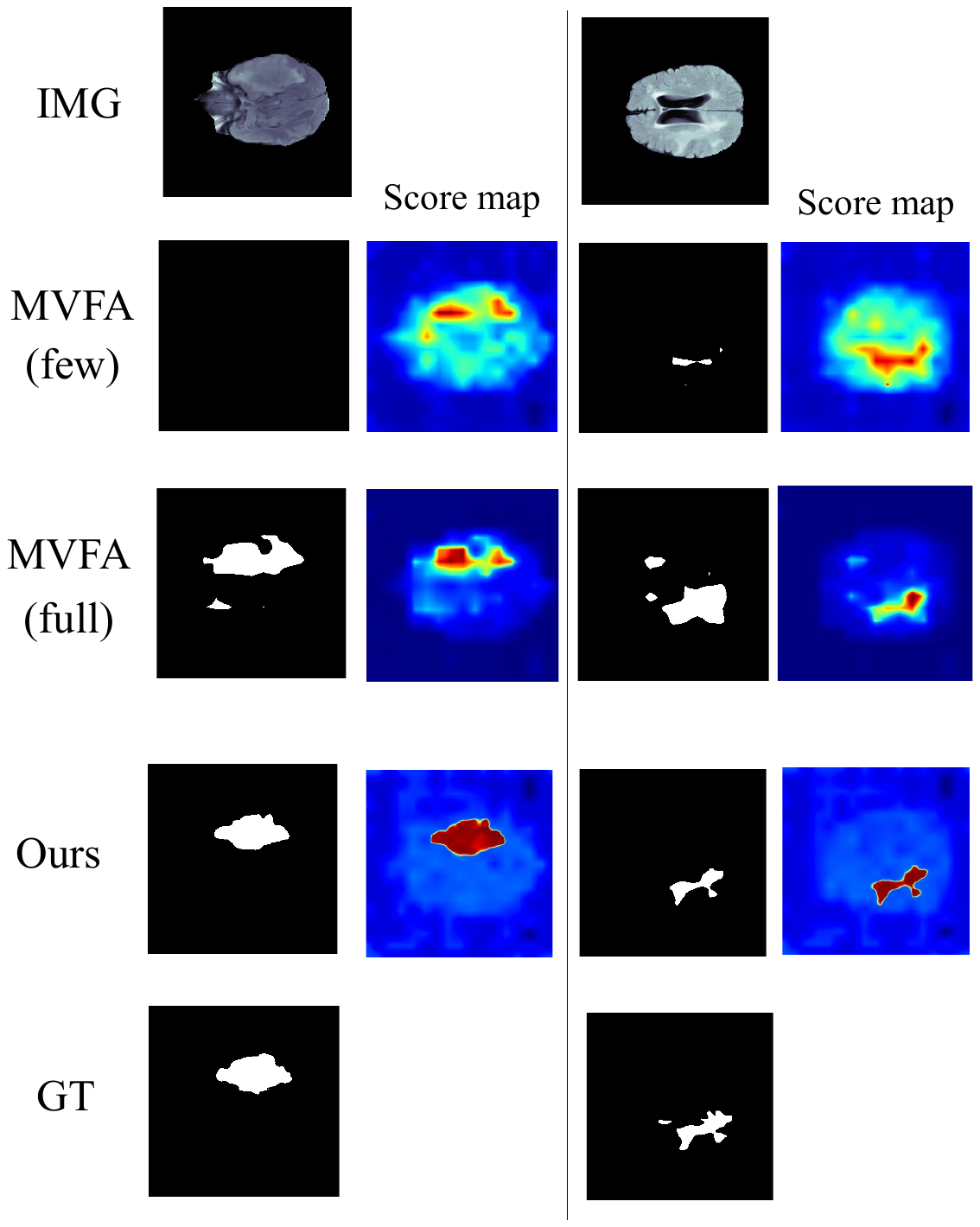}
  \caption{Qualitative results comparing MVFA (few/full) and our method on brain anomaly segmentation, showing both binary masks and score maps.}
  \label{fig:dice_auc_qualitative}
\end{figure*}

AUC (or pixel-level pAUC) is a widely used metric for anomaly detection because it evaluates the model’s ability to correctly rank abnormal and normal samples across all thresholds. It is particularly useful when assessing global discriminative power or when the primary goal is to verify whether a model can reliably distinguish between healthy and abnormal cases. However, AUC has notable limitations that it does not reflect spatial localization quality, it remains insensitive to boundary errors, and high AUC values can arise even when the predicted anomaly maps are coarse. In clinical workflows, this can be problematic, as doctors must still manually examine the entire image to identify the exact lesion, increasing review time despite strong AUC scores. In contrast, the Dice coefficient directly measures spatial overlap between predictions and ground-truth lesions, making it far more indicative of precise localization. High Dice scores correspond to accurate, well-defined lesion boundaries, which substantially reduces the burden on clinicians by minimizing the amount of manual inspection required during image interpretation. 

As shown in ~\cref{tab:dice_auc_quantitative}, both AdaCLIP\cite{cao2024adaclip} and MVFA\cite{huang2024adapting} frequently achieve very high pAUC scores, often exceeding 90\% and even reaching 99\% in several datasets. However, their Dice scores remain substantially lower, with many cases falling below 50\% and even dropping to single digits on challenging datasets such as Lung Infection. This discrepancy indicates that these models can correctly rank abnormal pixels above normal pixels on average, yet fail to produce spatially coherent lesion masks. In other words, pAUC reflects the model’s ability to highlight anomalous regions in a coarse, score-based manner, while the low Dice values reveal that such signals lack the precise localization needed for reliable segmentation. Consequently, although high pAUC may suggest good anomaly discrimination, the absence of accurate boundaries forces clinicians to visually re-identify lesion extents, limiting the practical usefulness of these methods in real diagnostic workflows.

The qualitative results in ~\cref{fig:dice_auc_qualitative} further illustrate this limitation. For recent zero/few-shot works \cite{cao2024adaclip, huang2024adapting, ma2025aa}, the predicted score maps often exhibit broad, diffused activations that roughly highlight anomalous areas but fail to correspond to the exact lesion geometry. These models tend to generate large blobs of high response or scattered noisy activations, indicating that they capture global anomaly cues but lack the spatial precision required for clean segmentation. As a result, although their score maps contain some degree of anomaly awareness, the predicted masks derived from these scores are either overly coarse or incomplete, with boundaries that deviate significantly from the ground truth. This behavior explains the consistently low Dice scores that the models can rank abnormal regions correctly, reflected in high pAUC, but cannot accurately delineate the lesion contours, leading to poor boundary quality and unreliable localization.


\section{Additional Ablation Study on Hyperparameter Settings}
\subsection{Length of Learnable Tokens and Margin for \mylossfull} 
\begin{table}[h]
\centering
\caption{Effect of the number of learnable tokens on Dice (\%) and Accuracy (\%) across four datasets.}
\label{tab:tokens}
\resizebox{\columnwidth}{!}{
\begin{tabular}{c|cc|cc|cc|cc}
\hline
\multirow{2}{*}{\textbf{\#Tokens}} 
& \multicolumn{2}{c|}{\textbf{Brain}} 
& \multicolumn{2}{c|}{\textbf{Retina}} 
& \multicolumn{2}{c|}{\textbf{Lung}} 
& \multicolumn{2}{c}{\textbf{Breast}} \\
& Dice & Acc & Dice & Acc & Dice & Acc & Dice & Acc \\
\hline
10 & \textbf{89.47} & \textbf{96.34} & \textbf{93.18} & 97.37 & \textbf{87.16} & \textbf{99.22} & \textbf{84.96} & \textbf{91.01} \\
20 & 86.37 & 93.90 & 89.68 & 97.36 & 86.71 & 97.65 & 75.28 & 89.88 \\
35 & 85.42 & 89.02 & 90.73 & \textbf{98.24} & 86.44 & 99.21 & 76.41 & 89.89 \\
\hline
\end{tabular}
}
\end{table}

\begin{table}[h]
\centering
\caption{Effect of different margins in \myloss on Dice (\%) and Accuracy (\%) across four datasets.}
\label{tab:margin_effect}
\resizebox{\columnwidth}{!}{
\begin{tabular}{c|cc|cc|cc|cc}
\hline
\multirow{2}{*}{\textbf{Margin}}
& \multicolumn{2}{c|}{\textbf{Brain}} 
& \multicolumn{2}{c|}{\textbf{Retina}} 
& \multicolumn{2}{c|}{\textbf{Lung}} 
& \multicolumn{2}{c}{\textbf{Breast}} \\
& Dice & Acc & Dice & Acc & Dice & Acc & Dice & Acc \\
\hline
0.2 & 84.87 & 89.02 & 92.18 & 97.36 & 84.64 & 98.43 & 76.40 & \textbf{94.38} \\
0.4 & \textbf{89.47} & \textbf{96.34} & \textbf{93.18} & \textbf{97.37} & \textbf{87.16} & \textbf{99.22} & \textbf{84.96} & 91.01 \\
0.6 & 84.66 & 93.90 & 89.69 & 96.49 & 86.21 & 98.45 & 76.38 & 90.44 \\
0.8 & 81.17 & 82.92 & 62.92 & 96.49 & 85.53 & 98.43 & 75.28 & 92.69 \\
\hline
\end{tabular}
}
\end{table}

~\cref{tab:tokens} and ~\cref{tab:margin_effect} present two key ablations on the number of learnable prompt tokens and the margin value in the margin-contrastive loss. For learnable tokens, using 10 tokens consistently achieves the best performance across Brain, Retina, and Lung, with the highest Dice scores (89.47\%, 93.18\%, and 87.16\%, respectively). Increasing the prompt length to 20 or 35 leads to clear degradation, indicating that excessive learnable tokens may introduce redundancy and degrade image-text alignment. For the contrastive margin, a margin of 0.4 provides the most stable and discriminative learning signal, producing the top Dice scores on Brain and Retina while avoiding the severe performance collapse observed at larger margins (e.g., Dice drops to 62.92\% on Retina when margin = 0.8). These results collectively indicate that a compact prompt representation (10 tokens) combined with a moderate contrastive separation (margin = 0.4) yields the most robust alignment and segmentation performance.

\subsection{Different Threshold for Dice score of Baselines}
\begin{table}[h]
\centering
\caption{Comparison of threshold-based Dice (\%) across baseline models on four datasets.}
\label{tab:threshold_comp}
\resizebox{\linewidth}{!}{
\begin{tabular}{c|cccc|cccc}
\hline
& \multicolumn{4}{c|}{\textbf{Aaclip}\cite{ma2025aa}} & \multicolumn{4}{c}{\textbf{Anomalyclip}\cite{zhou2023anomalyclip}} \\
\hline
\textbf{Threshold} & Brain & Retina & Lung & Breast & Brain & Retina & Lung & Breast \\
\hline
0.3 & 14.04 & 38.03 & 9.54  & 10.74 & 46.99 & 39.13 & 44.53 & 76.40 \\
0.4 & 18.59 & 46.94 & 12.06 & 20.81 & 46.99 & 39.13 & 44.53 & 76.40 \\
0.5 & 22.68 & 47.98 & 21.73 & 41.54 & 46.99 & 39.13 & 44.53 & 76.40 \\
0.6 & 35.48 & 43.94 & 27.32 & 56.92 & 46.99 & 39.13 & 44.53 & 76.40 \\
0.7 & 41.59 & 41.88 & 34.79 & 67.21 & 46.99 & 39.13 & 44.53 & 76.40 \\
\hline
& \multicolumn{4}{c|}{\textbf{MadCLIP}\cite{shiri2025madclip}} & \multicolumn{4}{c}{\textbf{Adaclip}\cite{cao2024adaclip}} \\
\hline
\textbf{Threshold} & Brain & Retina & Lung & Breast & Brain & Retina & Lung & Breast \\
\hline
0.3 & 0.34 & 0.78 & 0.70 & 0.35 & 42.08 & 39.05 & 6.45  & 32.54 \\
0.4 & 0.38 & 0.81 & 0.71 & 0.41 & 42.66 & 39.96 & 7.38  & 35.19 \\
0.5 & 0.38 & 0.82 & 0.67 & 0.43 & 43.52 & 38.62 & 9.13  & 36.19 \\
0.6 & 0.36 & 0.81 & 0.58 & 0.41 & 44.58 & 39.41 & 13.47 & 36.03 \\
0.7 & 0.33 & 0.78 & 0.45 & 0.37 & 45.78 & 39.63 & 18.03 & 36.89 \\
\hline
\end{tabular}
}
\end{table}

~\cref{tab:threshold_comp} presents an ablation study on threshold selection for converting raw anomaly score maps (ranging from 0 to 1) into binary segmentation masks for baseline models, ensuring a fair and objective comparison against our method. Since these raw models originally produced anomaly score with pixel value range from 0 to 1, and a decision threshold must be chosen to evaluate segmentation performance. While A default threshold of 0.5 is commonly used,  we vary thresholds from 0.3 to 0.7 to examine the model performance variance.  We do not use extremely low thresholds effectively classify nearly all pixels as anomalous (over-segmentation), whereas extremely high thresholds suppress most anomalies (under-segmentation). 

The results show that Anomalyclip is relatively insensitive to threshold variations, maintaining stable Dice scores across datasets. In contrast, models such as Aaclip, Adaclip exhibit noticeable fluctuations. For instance, Aaclip achieves its best Dice on Brain at threshold 0.7, yet performs worse at intermediate values, highlighting the instability of selecting a “best” threshold universally. Across the broader results, threshold 0.5 consistently provides either the highest or a competitive Dice score for most methods and datasets, without the extreme behavior observed at 0.3 or 0.7. In conclusion, 0.5 is adopted as a fair and balanced threshold for evaluating raw anomaly-score-based CLIP models, offering a stable compromise across models and datasets without favoring over- or under-segmentation.

\end{document}